\documentclass[runningheads]{llncs}
\usepackage{graphicx}

\usepackage{amsmath}
\usepackage{amssymb}
\usepackage{verbatim}
\usepackage{graphicx}

\usepackage{color}
\usepackage{xcolor}
\usepackage{soul}
\usepackage[normalem]{ulem}
\usepackage{authblk} 

\definecolor{mypink1}{rgb}{0.858, 0.188, 0.478}
\definecolor{deepskyblue}{rgb}{0.0, 0.74901960784, 1.0} 

\newcommand{\sarah}[1]{{\textcolor{mypink1}{[#1]}}}
\newcommand{\matt}[1]{{\textcolor{deepskyblue}{[#1]}}}

\newcommand{\PREV}[1]{{\textcolor{blue}{\textst{[#1]}}}}
\newcommand{\ek}[1]{{\textcolor{olive}{[#1]}}}

\begin{document}
\title{Uncertainty Quantification in CNN-Based Surface Prediction Using Shape Priors}
\titlerunning{Uncertainty Quantification in CNN-Based Surface Prediction}

\author{Katar\'{i}na T\'{o}thov\'{a}\inst{1}\and Sarah Parisot\inst{2} \and Matthew C. H. Lee\inst{3} \and Esther Puyol-Ant\'{o}n\inst{4} \and Lisa M. Koch\inst{1} \and Andrew P. King\inst{4} \and Ender Konukoglu\inst{1} \and Marc Pollefeys \inst{1,5}}
\authorrunning{K. T\'{o}thov\'{a} et al.}
%
\institute{ETH Zurich, Switzerland \and Aimbrain ltd. \and HeartFlow Inc. \and King's College London, United Kingdom \and Microsoft}
\maketitle              

\begin{abstract}
Surface reconstruction is a vital tool in a wide range of areas of medical image analysis and clinical research. Despite the fact that many methods have proposed solutions to the reconstruction problem, most, due to their deterministic nature, do not directly address the issue of quantifying uncertainty associated with their predictions. We remedy this by proposing a novel probabilistic deep learning approach capable of simultaneous surface reconstruction and associated uncertainty prediction. The method incorporates prior shape information in the form of a principal component analysis (PCA) model. Experiments using the UK Biobank data show that our probabilistic approach outperforms an analogous deterministic PCA-based method in the task of 2D organ delineation and quantifies uncertainty by formulating distributions over predicted surface vertex positions.

\end{abstract}


\section{Introduction}
Reconstruction of organ surfaces and segmentation of their bodies are 
amongst the most important tasks in medical image analysis. High-quality organ surface models are often sought after in disciplines such as cardiac or neuro-imaging, and provide a powerful tool in diagnosis, surgical planning, disease tracking, longitudinal studies and interpretation of functional data \cite{peressutti2017,puyol2017,bai2017}. 

Traditional approaches to parametric surface modelling rely on evolving deformable shapes according to predefined forces \cite{fischl2012,huo2016,han2004,schuh2017} or use atlas registration \cite{puyol2017,peressutti2017,garcia-barnes2010,bai2013}. Recent advances in machine learning have showed the possibility of training a deep neural network in an end-to-end manner: from images to parametrised shapes \cite{milletari2017}. In their work, Milletari et al. \cite{milletari2017} devised a convolutional neural network (CNN) to directly predict coordinates of the organ surface mesh from the imaging data. To build in prior shape knowledge, the network contains an explicit principal component analysis (PCA) layer and predictions are made as linear combinations of the modes of variation determined from the training data. 


Despite the advances in organ surface modelling and associated segmentation, ways of estimating the precision of the prediction are still sparse. Yet, it is of vital importance for interpreting medical data to be able to not only measure the accuracy of the result as a deterministic sample but to quantify the uncertainty associated with it as well. We can distinguish between two types of uncertainty \cite{kendall2017}: \textit{aleatoric} uncertainty inherent to the data, modelled by probability distribution over model outputs, and \textit{epistemic} uncertainty accounting for uncertainty in the model parameters, which typically decreases with increasing data size.

In the context of medical imaging, the need for addressing aleatoric uncertainty stems from the nature of the data. Medical imaging data often suffers from high levels of noise, coarse resolution and imaging artifacts - all the factors conspiring towards heightened need for quantification of uncertainty about the produced results. In image segmentation, this has been approached by means of segmentation sampling \cite{le2016,chang2011}, where several  plausible segmentations are gathered to estimate the variability of the output. While \cite{chang2011} uses MCMC to sample segmentations from an estimated posterior distribution where likelihood and prior functions are defined, \cite{le2016} introduces Gaussian processes to sample from the posterior directly, knowing only its mean and covariance. On the other hand, uncertainties inherent to the prediction models can be captured by means of distributions over model parameters. Bayesian neural networks \cite{denker1991,mackay1992,neal1995,gal2016}, where one puts priors over the model parameters instead of using deterministic values, have been employed to this end. Extensions combining both aleatoric and epistemic uncertainty into one model have been proposed in \cite{kendall2017,kwon2018}.

In our work, we build on a PCA-based method of surface reconstruction \cite{milletari2017} and propose a probabilistic approach to integrate aleatoric uncertainty quantification within the model. We formulate the problem as a conditional probability estimation incorporating shape information in the form of a PCA model.

Hence, our approach addresses three main objectives: 
\begin{itemize}
	\item Direct probabilistic surface mesh prediction from imaging data. The proposed method improves upon a deterministic direct coordinate prediction by up to 12\% on the UK BioBank dataset \cite{UKBB} as measured by DICE.
    \item Use of PCA-based shape priors to predict sensible shapes only. Here, our probabilistic approach improves upon the deterministic PCA-based method by up to 10.7\% in terms of DICE score. 
	\item Novel aleatoric uncertainty quantification formulation by means of assessing the posterior of the predicted surface.
\end{itemize}

\section{Prediction Model}
We formulate the surface prediction problem via a probabilistic model that utilises principles from probabilistic PCA \cite{bishop}.
Our goal is to build a model that goes beyond deterministic prediction and allows sampling 2D delineations of the organ surface based on the corresponding MRI data. 
To this end, we express the prediction as a probability conditioned on the image, $p(y|x)$, where $x$ refers to the MRI image and $y$ to the parametrised surface, i.e. set of surface vertices. 
We model the conditional probability $p(y|x)$ with a latent variable model
\begin{equation}\label{pyx}
    p(y|x) = \int p(y|z,x) p(z|x) dz,
\end{equation}
where $z$ is the set of latent variables. 

The PCA aspect of the model lies within the definition of $p(y|z,x)$
\begin{equation}\label{pyz}
    p(y|z,x) = \mathcal{N} ( y | U S^{\frac{1}{2}}z + \mu + s(x), {\sigma}^2 I),
\end{equation}
Here, $U$, $\mu$ 
and $S$ are the principal component matrix (principal components are columns of the matrix), mean and diagonal covariance matrix respectively, all precomputed using the surfaces in the training set, and $s$ is a global spatial shift that depends on the image $x$. In this formulation, the latent variable $z$ can be interpreted as the PCA weights corresponding to the surface $y$. Variance $\sigma^2$ refers to the noise level in the data.

The conditioning to the image is modelled with $p(z|x)$ and we use a deep CNN architecture for this purpose. Specifically, we express
\begin{equation}\label{pzx}
	p(z|x) = \mathcal{N}(z|\mu(x), \Sigma(x)).
\end{equation}
Here, a deep network takes the image $x$ as input and predicts $\mu(x)$ and $\Sigma(x)$ simultaneously. 

The last component of the proposed model is the prior for the latent variables $p(z)$. The probabilistic PCA model assumes a unit Gaussian distribution for the PCA weights, i.e. 
$p(z)\sim \mathcal{N}(0,I)$. 
In our model we assume the same, which becomes important when training the model. 

\subsubsection*{Training}
During training we optimise the parameters of $\mu(x)$, $\Sigma(x)$ and $s(x)$ using a training set.  The optimisation objective consists of two terms: the first one aims to maximise the conditional probability $p(y|x)$ while the latter regularises the loss by minimising the Kullback-Leibler divergence (KLD) between the assumed prior distribution of $z$ and the observed one in the training set, i.e. $\int p(z|x)p(x)dx \approx \sum_x p(z|x)$. 

Direct maximisation of Equation~\ref{pyx} requires marginalisation of the latent variable. The marginal distribution is also Gaussian with analytical mean and variance that allow direct computation and therefore, optimisation of $\ln p(y|x)$. However, this requires inverting a not-necessarily-diagonal covariance matrix of the size $\textsf{(num of vertices)}^2$ at each iteration, which can be infeasible due to the size and potential numerical instabilities, as we empirically observed. Therefore, instead of directly maximising $\ln p(y|x)$, we use Jensen's inequality to derive a lower bound as follows
\begin{align}
    \ln p(y|x) \geq \mathbb{E}_{z|x}\left[ \ln p(y|z) \right] 
                                                \cong \frac{1}{L} \sum_{l=1}^L \ln p(y|z_l), \label{post}
\end{align}
where $z_l$ is sampled from $p(z|x)$ defined in (\ref{pzx}). 

Maximisation of the lower bound in Equation~\ref{post} will not necessarily satisfy the prior in the probabilistic PCA model. To address this, we use a regularisation term that aims to align the observed latent variable distribution with its prior. In order to satisfy the PCA model
\begin{align}
    p(z) \cong \sum p(z|x_n), \qquad x_n \sim p(x).
\end{align}
We use the KLD as a measure of deviation from this criteria and minimise 
\begin{equation}\label{KLD}
	\mathrm{KLD} \left( \sum_{n=1}^{N} p(z|x_n), p(z) \right)
\end{equation}

Using (\ref{post}) and (\ref{KLD}), our full model can then be trained by solving the following minimisation problem:
\begin{equation}\label{loss}
    \min_{\theta} \left\{\lambda \, \mathrm{KLD} \left( \sum_{n=1}^{N} p(z|x_n), p(z) \right) - \sum_{n=1}^N \frac{1}{L} \sum_{l=1}^L \ln p(y|z_l)\right\},  
\end{equation}
where $z_l\sim \mathcal{N}(z|\mu(x_n), \Sigma(x_n))$ and $\lambda$ is a regularisation parameter. 
For further details on the derivations please refer to Appendix A.

\subsubsection*{Inference}
As we mentioned previously, in the proposed model $p(y|x)$ is a Gaussian distribution. For a given test image, we perform prediction by computing the mean and the covariance matrix of this distribution.

Given (\ref{pyz}) and (\ref{pzx}) we can write
\begin{align}
	\mathbb{E}(y|x) & = U S^{\frac{1}{2}} \mu(x) + \mu + s, \\
    \mathrm{var}(y|x) & = \sigma^2I + U S^{\frac{1}{2}} \Sigma(x) (US^{\frac{1}{2}})^T.
\end{align}
Note that the first term of the predicted variance, i.e. $\sigma^2 I$, represents a fixed noise level in the data. Full derivation can be found in Appendix B.

\section{Experiments and Results}
We have tested the proposed method on a task of delineation of myocardium boundaries in cardiac MRI using imaging volumes from UK BioBank \cite{UKBB}. The~2D surface reconstruction consisting of the prediction of 50 vertices was evaluated on small ($60 \times 60$ crops around the heart) and full ($200 \times 200$ crops) field of view (FOV) images with the following characteristics: 
\begin{itemize}
	\item Small FOV: isotropic pixel of size 1.8\,mm; 572 training, 160 testing and 195 validation examples
    \item Full FOV: isotropic pixel of  size 1.8269\,mm; 1532 training, 455 testing and 499 validation examples
\end{itemize}
We used one imaging slice per original MRI volume. Active contours \cite{kass1988} delineations consisting of 50 connected (corresponding throughout the subjects) vertices were obtained from reference segmentations extracted from the UK BioBank dataset. These were prepared automatically using expert-segmentations and combination of learning and registration methods described in \cite{bai2017b,sinclair2017}.

The deep network architecture used in our model is analogous to the main branch described in \cite{milletari2017} with 9 convolutional, 3 pooling and one dense layer (CL9P3DL1). Convolutional layers were followed by ReLU activations. Training was done by minimising loss (\ref{loss}) using RMS-Prop optimiser with a constant learning rate $10^{-6}$ for batches of size 5. Noise level in the data $\sigma^2$ and regularisation parameter $\lambda$ were empirically set to $\sigma^2 = 5\times 10^{-2}$ and $\lambda = 10^{5}$.  

Table \ref{tabDICE} shows evaluation of segmentations obtained from myocardium delineation in terms of DICE score measuring region overlap and Root Mean Square Error (RMSE) comparing distances between the corresponding predicted and reference vertices. We used two different methods as baselines for comparison. Firstly, a network with the same architecture as the one employed in our model (CL9P3DL1) was used to directly predict the coordinates of the surface vertices (Direct Vertex). Secondly, we utilised the deterministic PCA approach (detPCA) based on \cite{milletari2017} again following the same architecture (without the spatial transformer refinement). For our method, the mean prediction $\mathbb{E}(y|x)$ served for computation of DICE scores and RMSE. 

\begin{table}[t]
\caption{\textbf{Vertex coordinate regression results.} Comparison of direct vertex prediction (Direct Vertex), deterministic PCA (detPCA) and our proposed probabilistic PCA-based approach (probPCA) with varying number of principal components. Segmentations used for computation of DICE were obtained by flooding the corresponding delineations. RMSE directly compares the predicted vertex coordinates to the reference ones.}
\centering
\begin{tabular}{|l|c|c|}
\hline
\textbf{Small FOV}&  DICE & RMSE \\
\hline
Direct Vertex & 0.86 $\pm$ 0.06 & 2.21 $\pm$ 1.04\\
detPCA 12& 0.87 $\pm$ 0.07 & 2.42 $\pm$ 1.27 \\
detPCA 8& 0.84 $\pm$ 0.07 & 2.51 $\pm$ 1.26 \\ \hline 
\multicolumn{3}{ c }{\textit{Ours:}} \\ \hline
probPCA 12 & \textbf{0.88 $\pm$ 0.09} & 1.91 $\pm$ 1.05 \\
probPCA 8  & \textbf{0.88 $\pm$ 0.09} & \textbf{1.80 $\pm$ 0.97} \\
\hline
\end{tabular}
\quad
\begin{tabular}{|l|c|c|}
\hline
 \textbf{Full FOV} &  DICE & RMSE \\
\hline
Direct Vertex & 0.75 $\pm$ 0.14 &  4.20 $\pm$ 2.82\\
detPCA 12& 0.78 $\pm$ 0.13  & 4.20 $\pm$ 2.65 \\
detPCA 8 & 0.75 $\pm$ 0.14 & 4.48 $\pm$ 2.69 \\ \hline
\multicolumn{3}{ c }{\textit{Ours:}} \\ \hline
probPCA 12 & \textbf{0.84 $\pm$ 0.10} & \textbf{2.59 $\pm$ 2.37}\\
probPCA 8  & 0.84 $\pm$ 0.11 & 2.60 $\pm$ 2.38\\
\hline
\end{tabular}
\qquad
\label{tabDICE}
\end{table}

\begin{figure}
\begin{minipage}[c]{0.325\textwidth}
	\includegraphics[angle=270,origin=c, width=\textwidth]{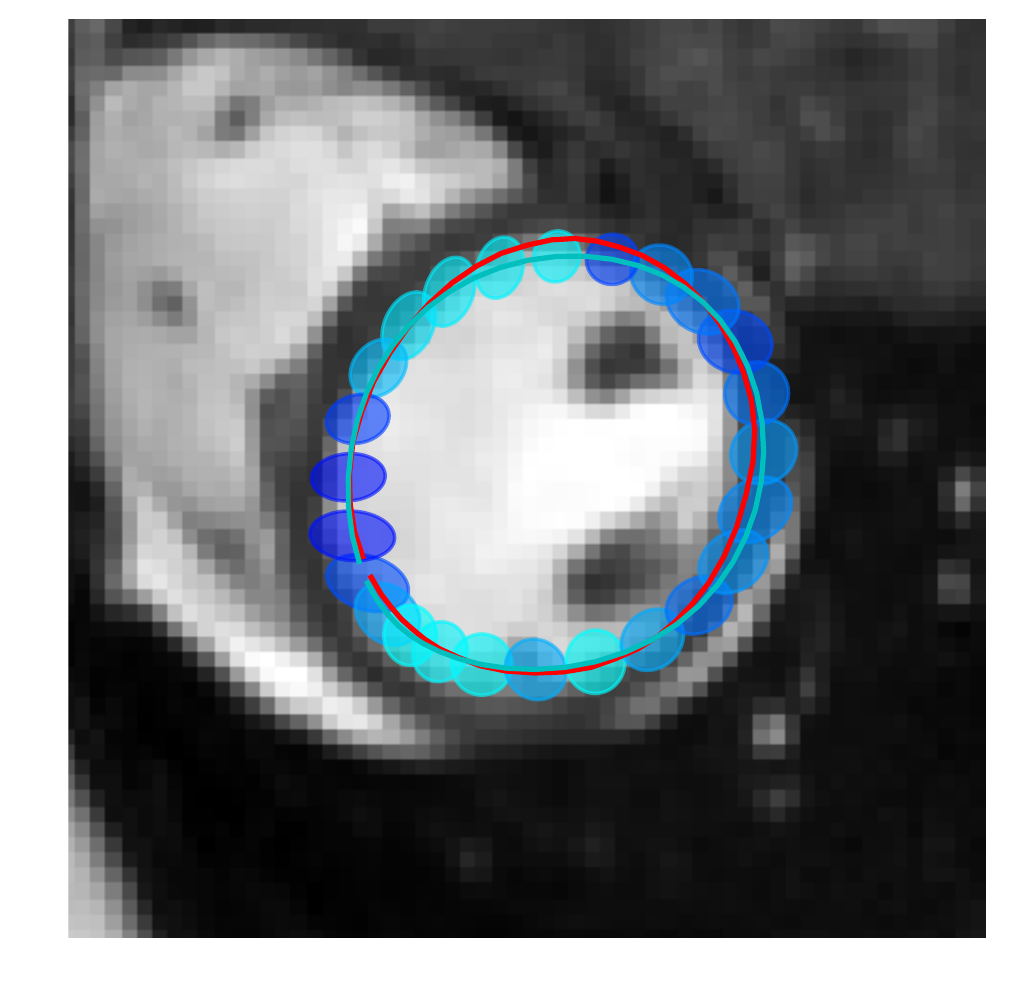}
\end{minipage}
\begin{minipage}[c]{0.325\textwidth}
	\includegraphics[angle=270,origin=c, width=\textwidth]{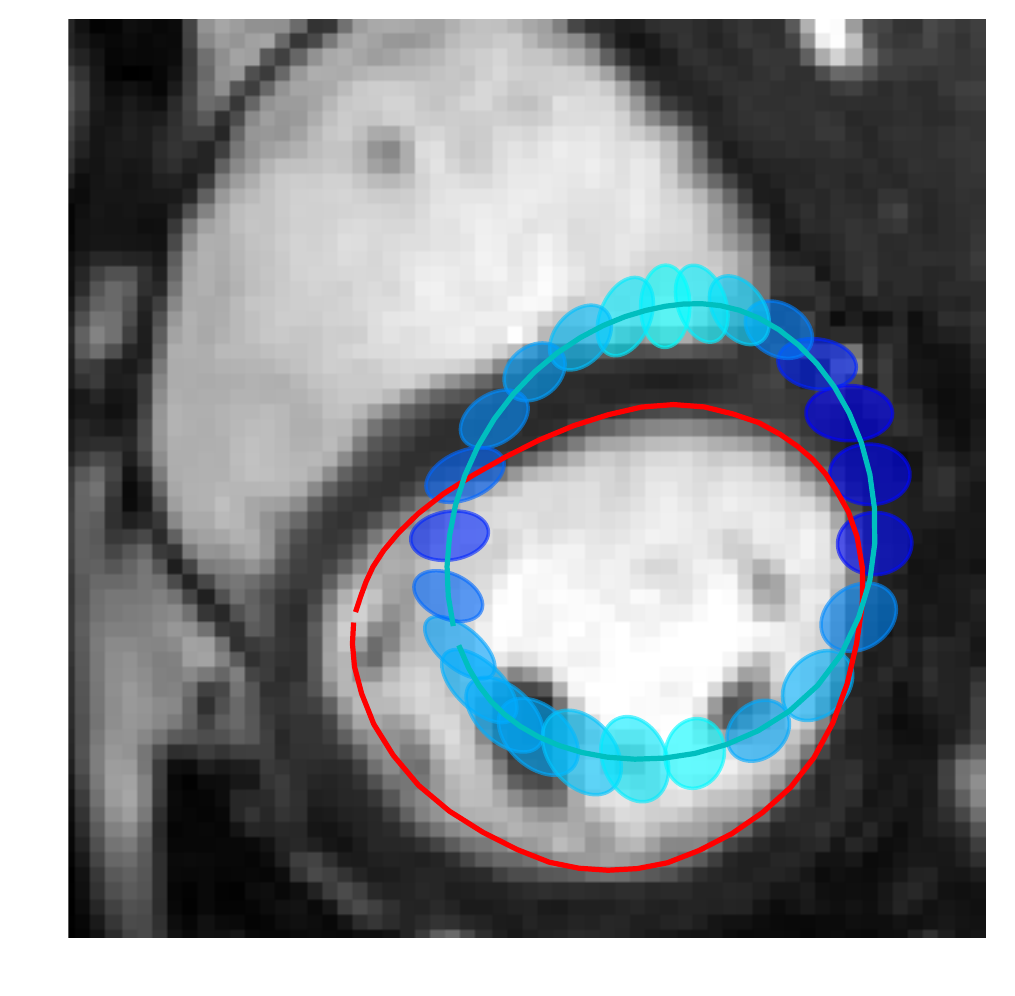}
\end{minipage}
\begin{minipage}[c]{0.325\textwidth}
	\includegraphics[angle=270,origin=c, width=\textwidth]{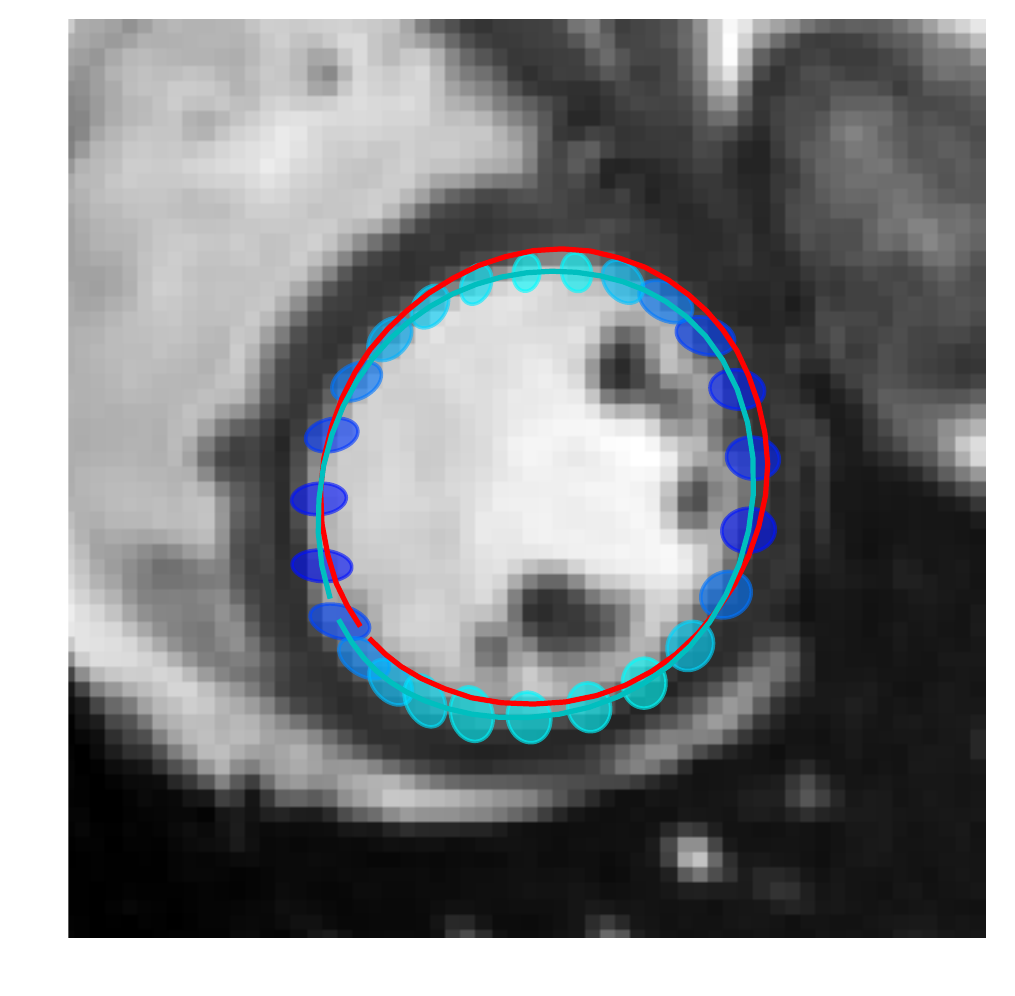}
\end{minipage}
\\
\begin{minipage}[c]{0.325\textwidth}
	\includegraphics[angle=270,origin=c, width=\textwidth]{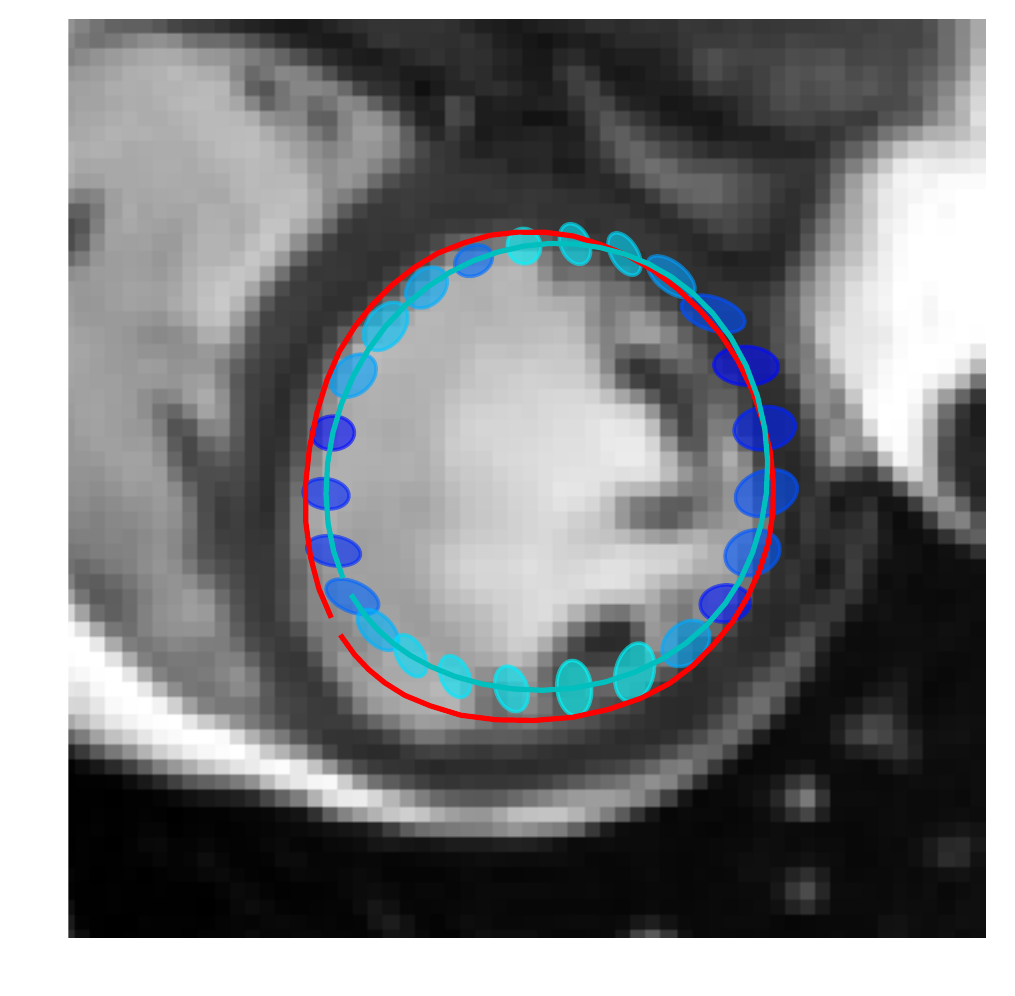}
\end{minipage}
\begin{minipage}[c]{0.325\textwidth}
	\includegraphics[angle=270,origin=c, width=\textwidth]{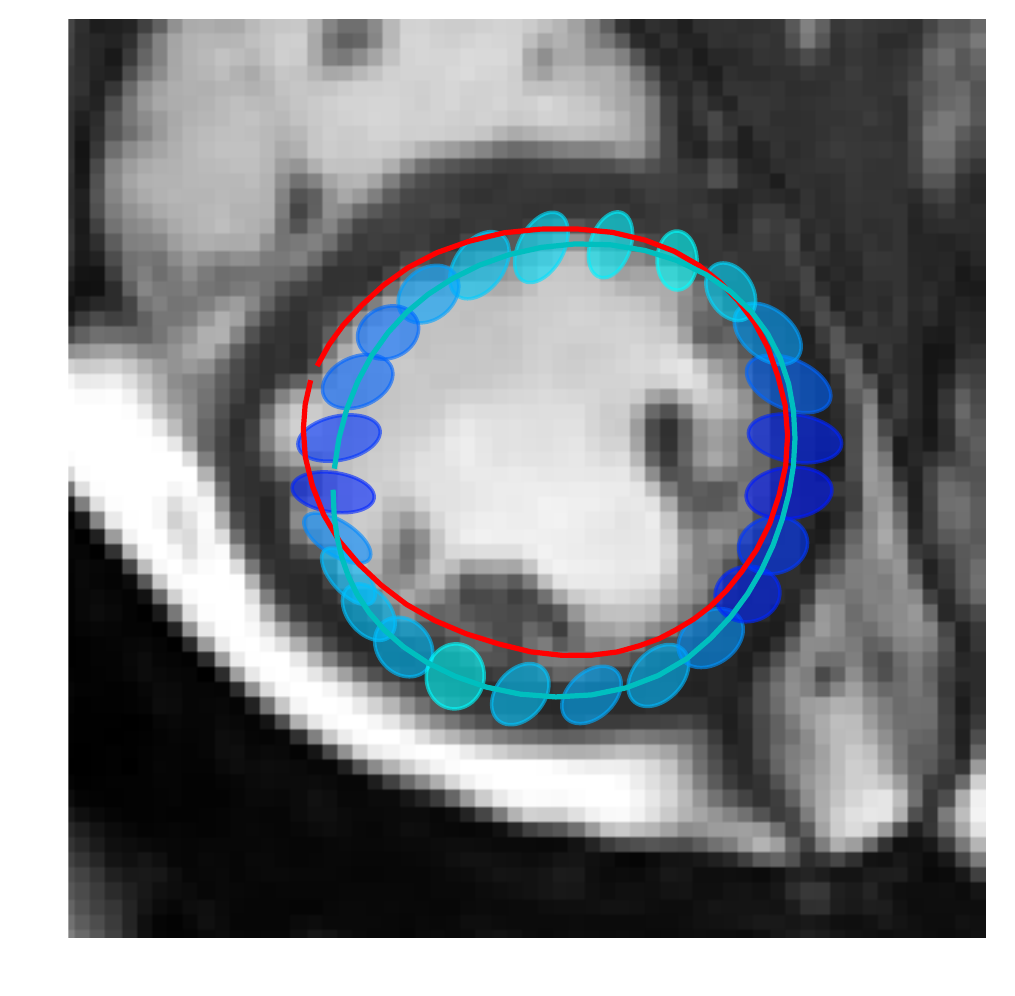}
\end{minipage}
\begin{minipage}[c]{0.325\textwidth}
	\includegraphics[angle=270,origin=c, width=\textwidth]{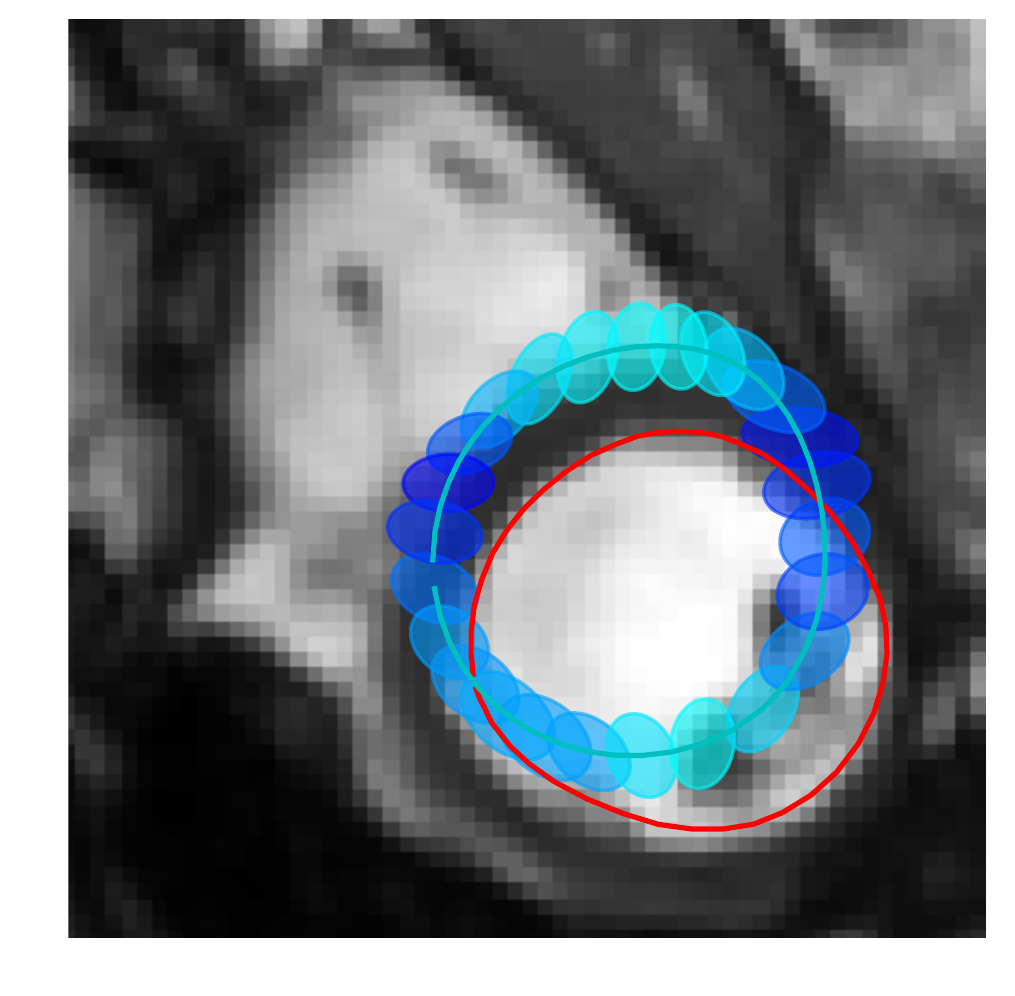}
\end{minipage}
\\
\begin{minipage}[c]{0.325\textwidth}
	\includegraphics[angle=270,origin=c, width=\textwidth]{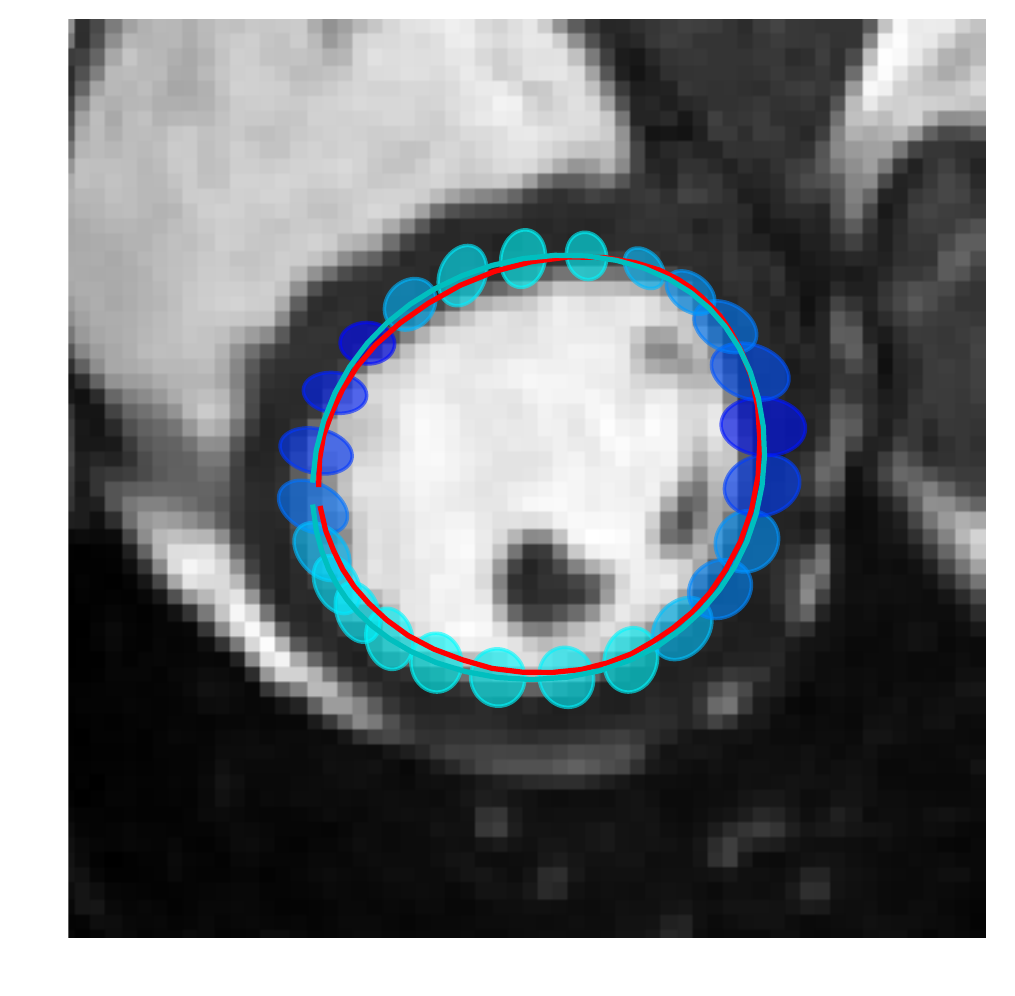}
\end{minipage}
\begin{minipage}[c]{0.325\textwidth}
	\includegraphics[angle=270,origin=c, width=\textwidth]{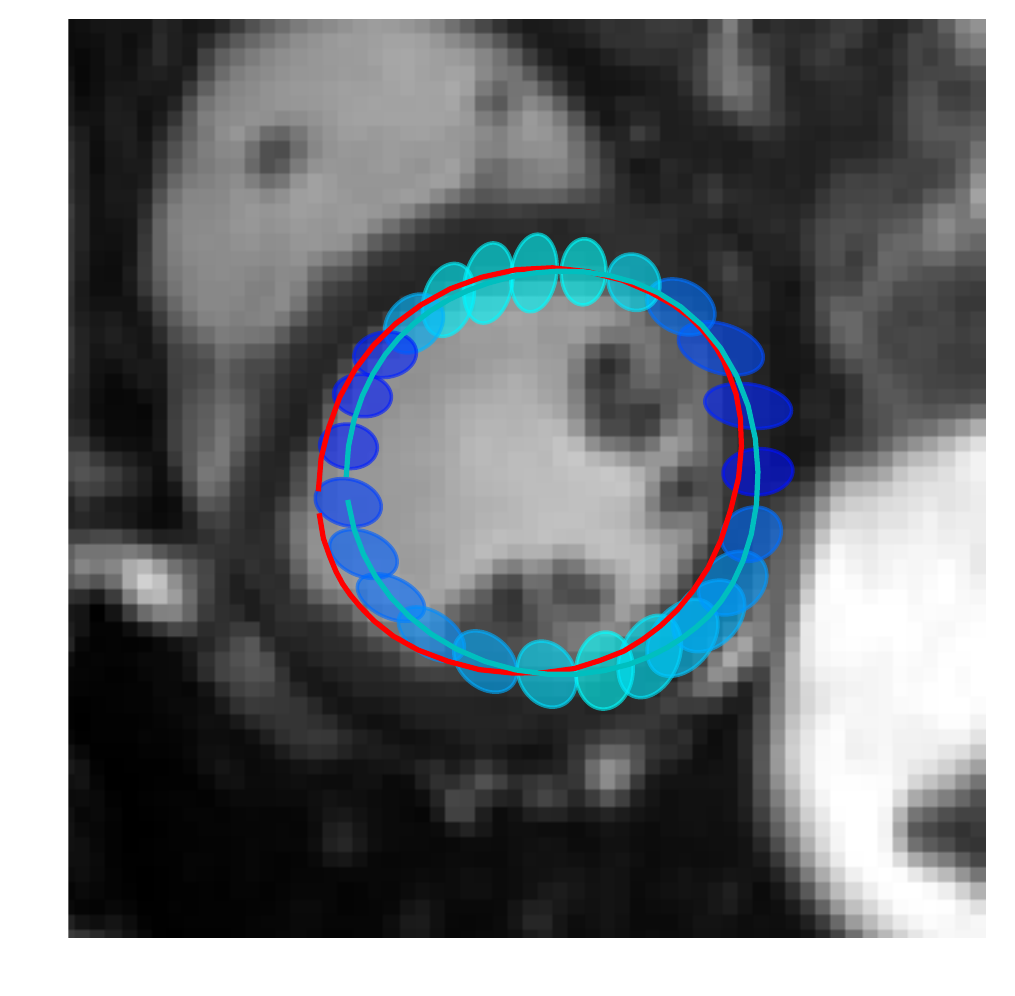}
\end{minipage}
\begin{minipage}[c]{0.325\textwidth}
	\includegraphics[angle=270,origin=c, width=\textwidth]{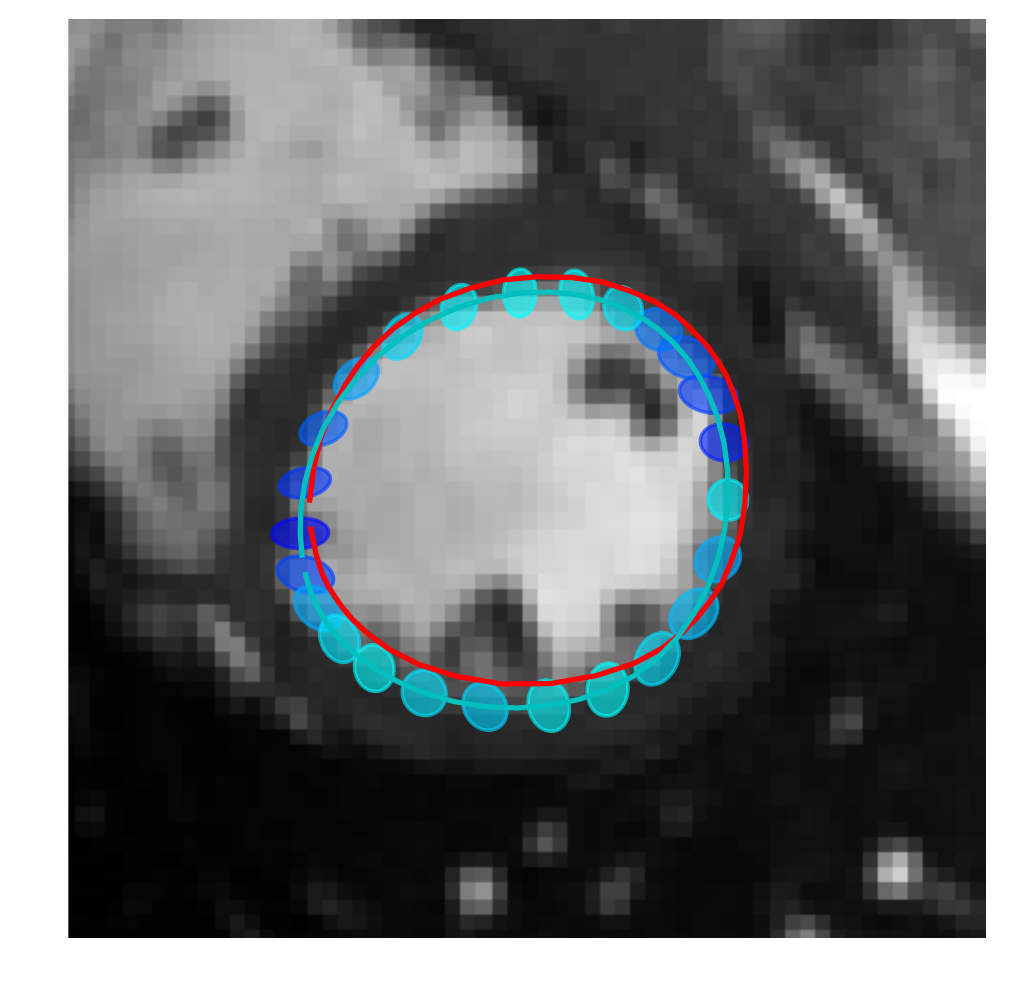}
\end{minipage}
\\
\begin{minipage}[c]{0.325\textwidth}
	\includegraphics[angle=270,origin=c, width=\textwidth]{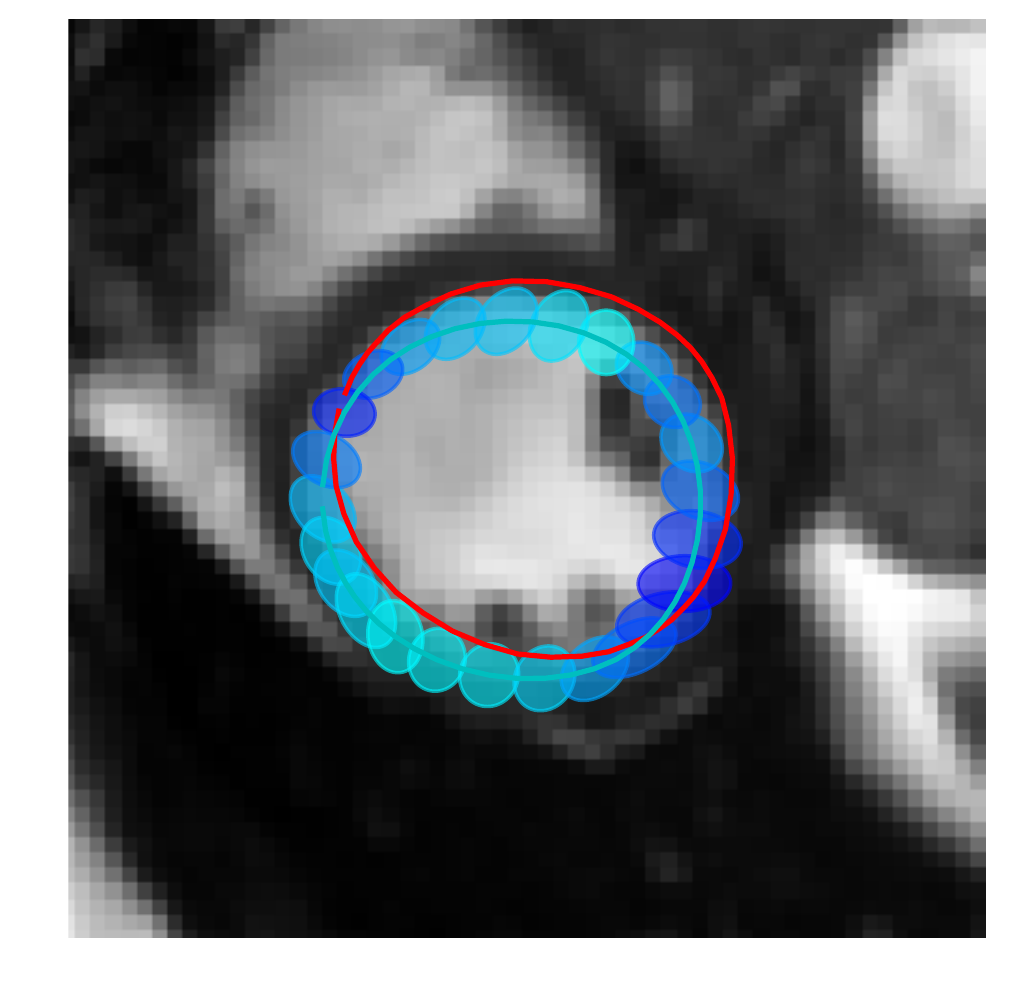}
\end{minipage}
\begin{minipage}[c]{0.325\textwidth}
	\includegraphics[angle=270,origin=c, width=\textwidth]{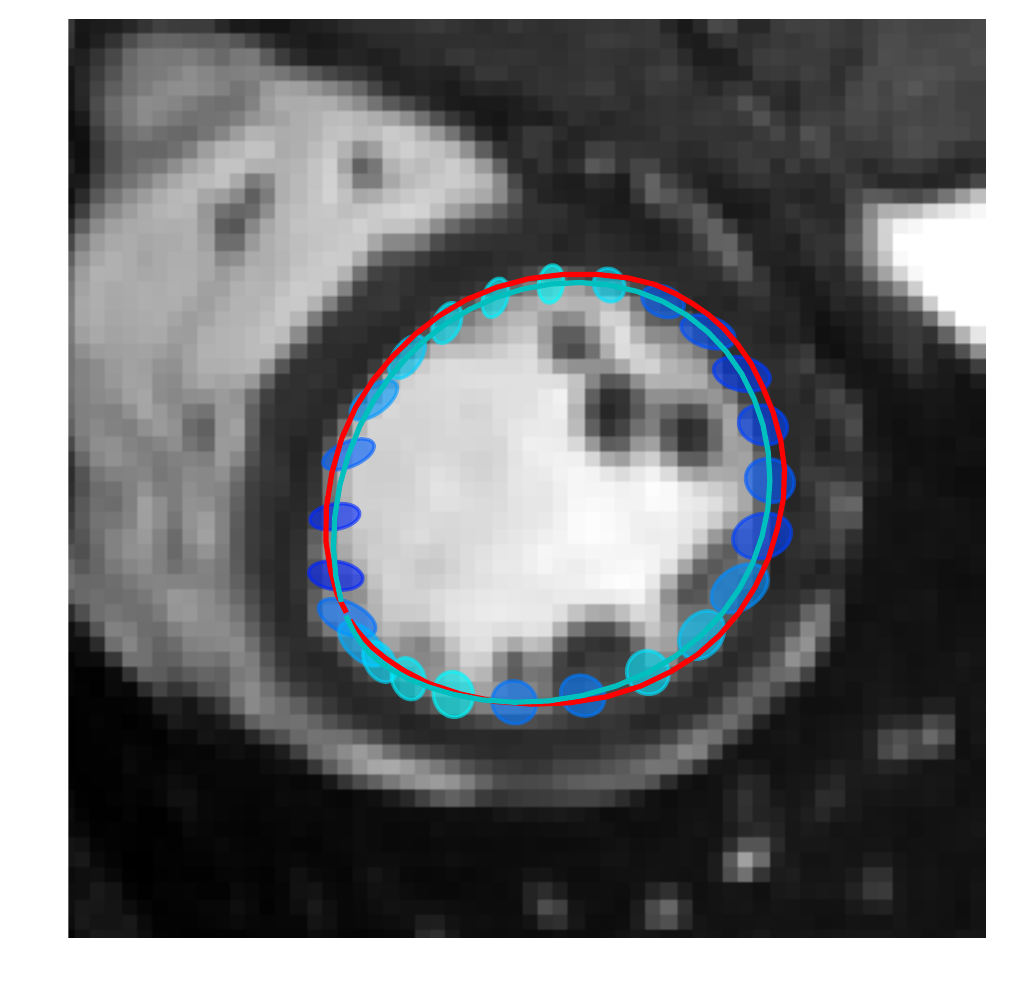}
\end{minipage}
\begin{minipage}[c]{0.325\textwidth}
	\includegraphics[angle=270,origin=c, width=\textwidth]{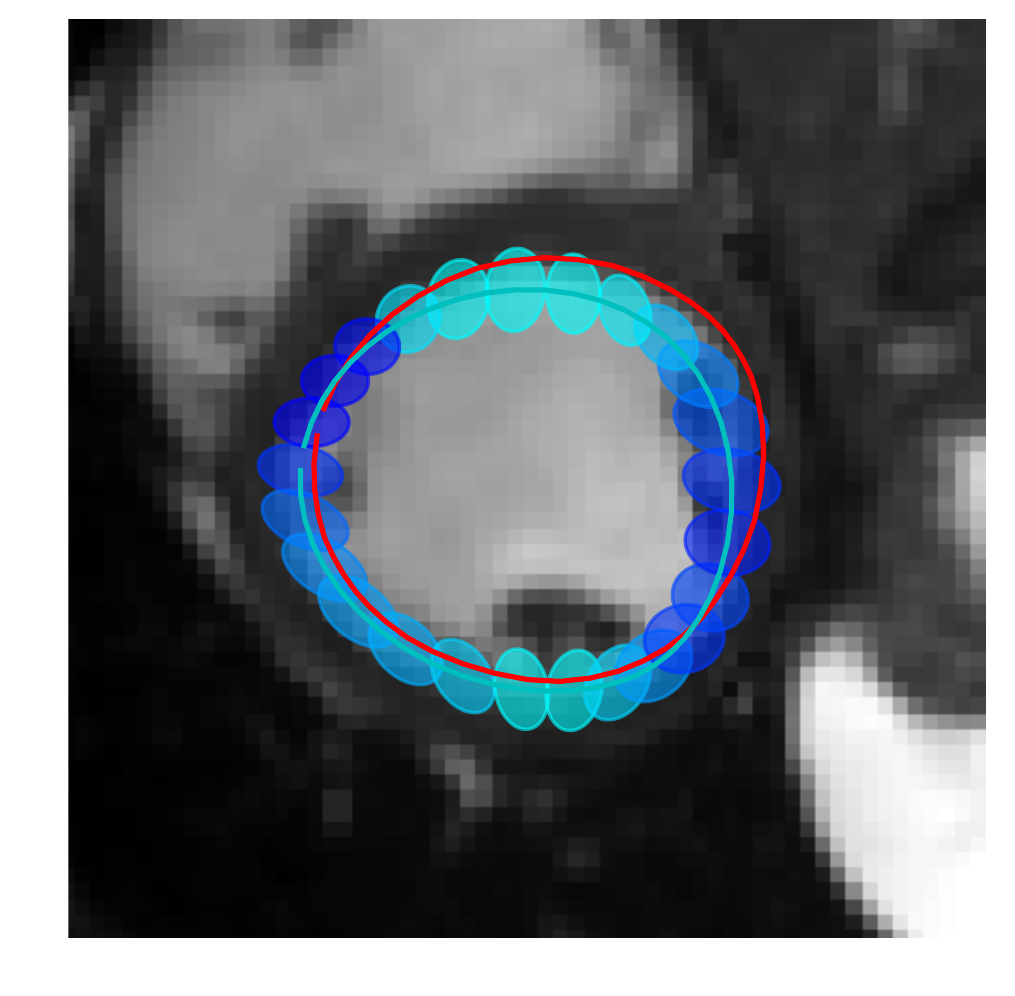}
\end{minipage}
\caption{Uncertainty quantification in small FOV myocardium delineations using 12 principal components at data noise level $5\times 10^{-2}$: 30\% confidence ellipses for points along the mean prediction. Colours correspond to the direction of the major axes. Only half of the points were plotted for clarity.}\label{glyphs}
\end{figure}

Working on the small FOV dataset, we exemplify the qualitative results of our probabilistic method in Figure \ref{glyphs}. Here, $p(y|x)$ distributions of predicted vertices are illustrated by plotting their variance as 30\% confidence ellipses along the predicted mean delineation. 
We only plot the 30\% confidence intervals and show results for small FOV for visualization purposes.
Notice how direction and size of the ellipses vary along the perimeter of the shape - the larger the variance the bigger the uncertainty over the position of the vertex that can be sampled from this distribution.

\begin{figure}[t]
\centering
\begin{minipage}[c]{0.325\textwidth}
	\includegraphics[angle=270,origin=c, width=\textwidth]{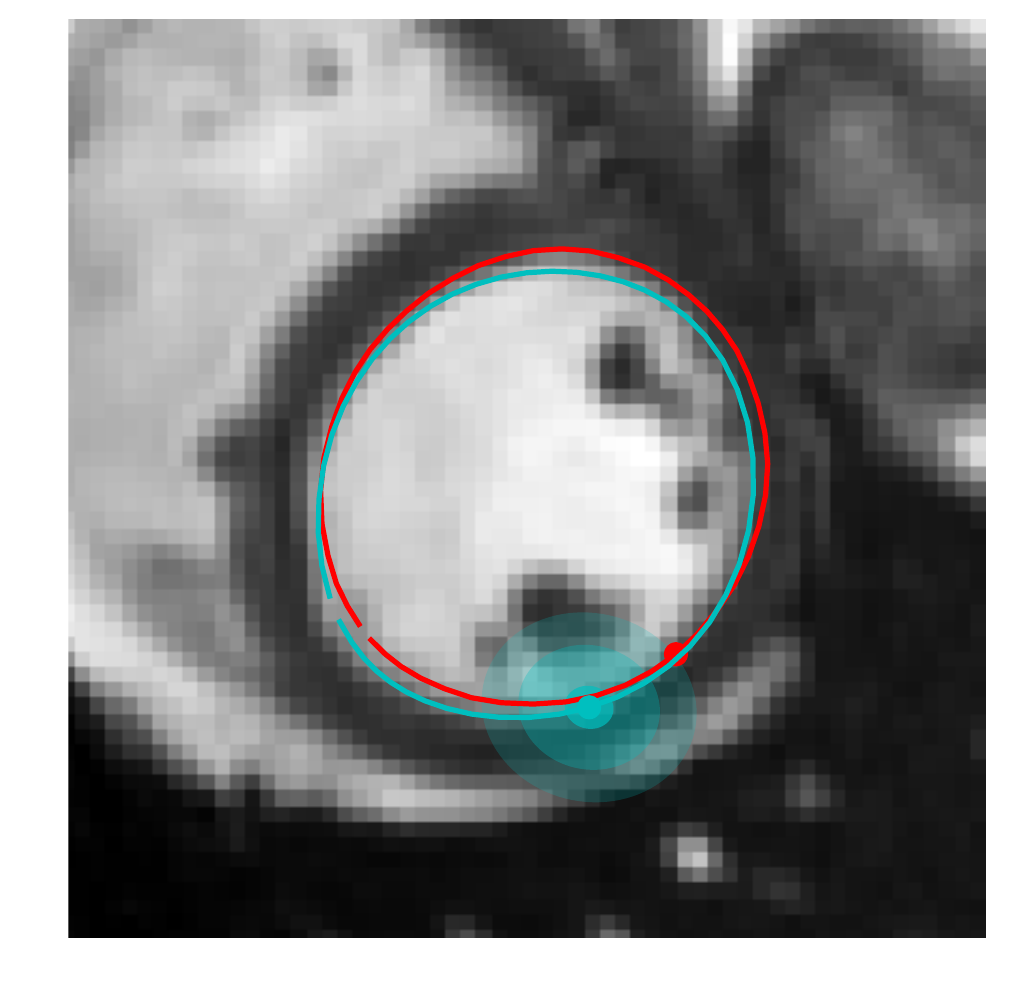}
\end{minipage}
\begin{minipage}[c]{0.325\textwidth}
	\includegraphics*[angle=270,origin=c, width=\textwidth]{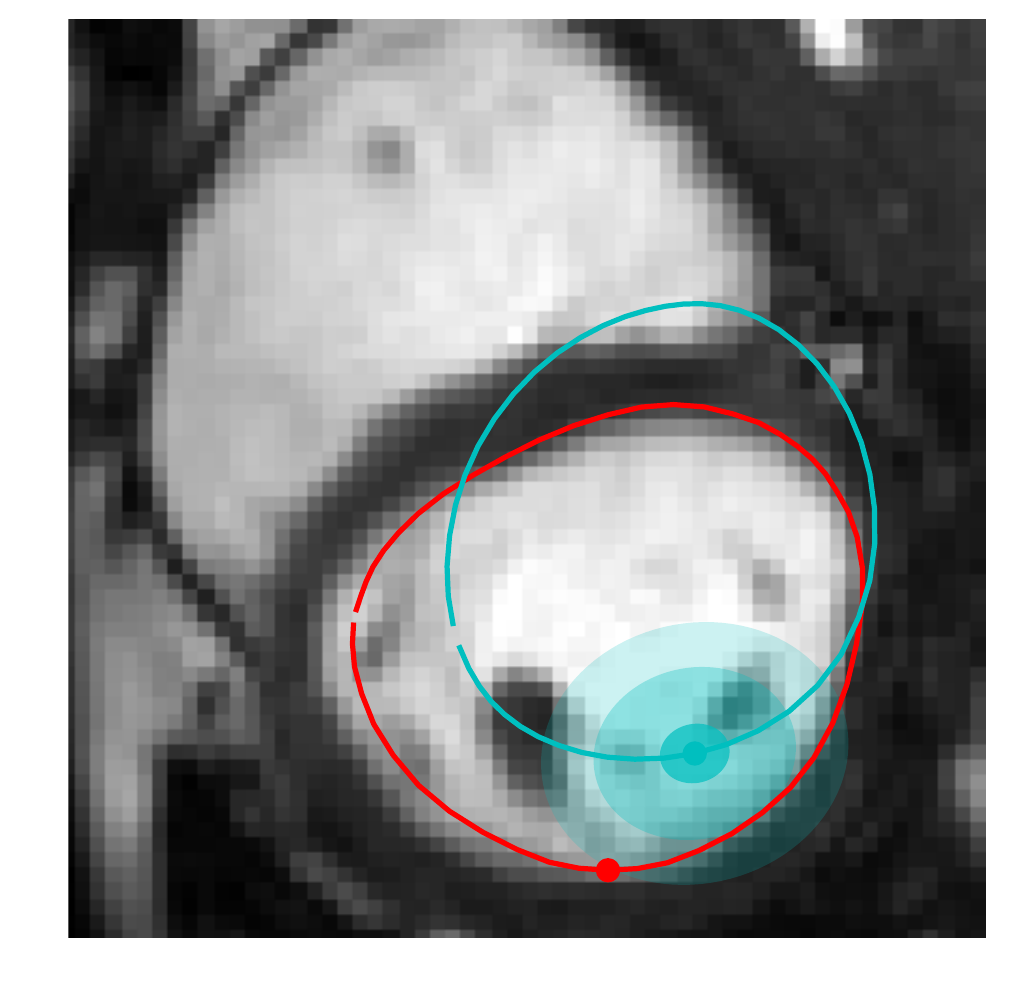}
\end{minipage}
\begin{minipage}[c]{0.325\textwidth}
	\includegraphics[angle=270,origin=c,  width=\textwidth]{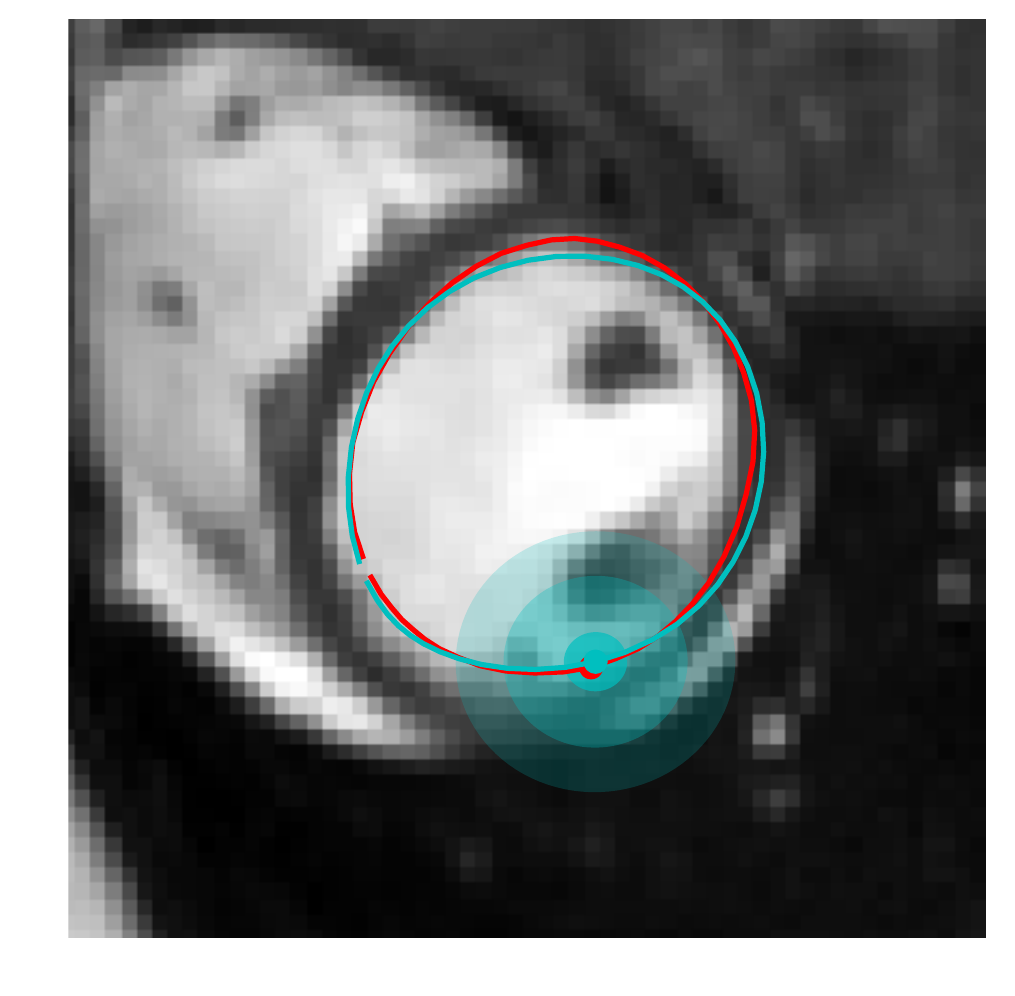}
\end{minipage}
\caption{Uncertainty quantification in small FOV myocardium delineation using 12 principal components at data noise level $5\times 10^{-2}$: reference (red), predicted mean(cyan). Posterior distribution of specific points from predictions above (reference point and its mean prediction marked with a dot). Ellipses correspond to 30, 95 and 99.9\% confidence regions. Observe how the reference point lies within the 99.9\% area. Note we used corresponding vertices throughout the subjects.} \label{smallFOV}
\end{figure}

Figure \ref{smallFOV} looks into the distribution of a specific vertex position in more detail. It essentially shows three types of results. On the left, we have high region overlap with high DICE (0.93) score, and relatively small variance; however large RMSE (3.23), which  can lead to problems if one was to use the retrieved surface for e.g. registration purposes. The middle figure is a failure case with high variance, low DICE(0.66) and high RMSE (6.12). And finally on the right is a result with high DICE (0.92) and low RMSE (0.83), but higher uncertainty than in the first case. Several conclusions can be drawn from this. Firstly, failure to predict the correct delineation leads to heightened uncertainty about the prediction. Secondly, what may seem as a good solution in terms of overlap (and hence DICE) may not necessarily be ideal in terms of RMSE. While delineations may align, the corresponding vertices may not. Lastly, even if the corresponding points do not align, the uncertainty over their position may still be relatively low (compare images on the left and right in Figure \ref{smallFOV}).  

\begin{figure}[t]
\centering
%
\begin{minipage}[c]{0.325\textwidth}
	\includegraphics[angle=270,origin=c, width=\textwidth]{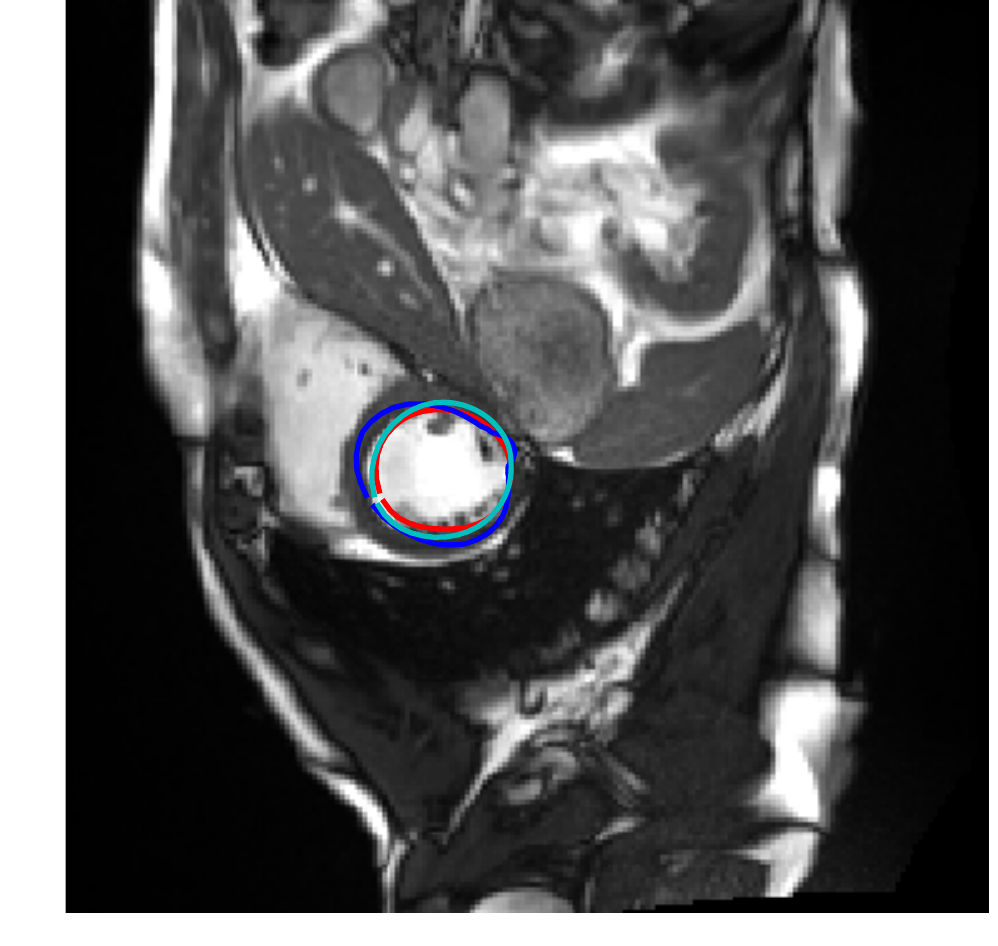}
\end{minipage}
\begin{minipage}[c]{0.325\textwidth}
	\includegraphics[angle=270,origin=c, width=\textwidth]{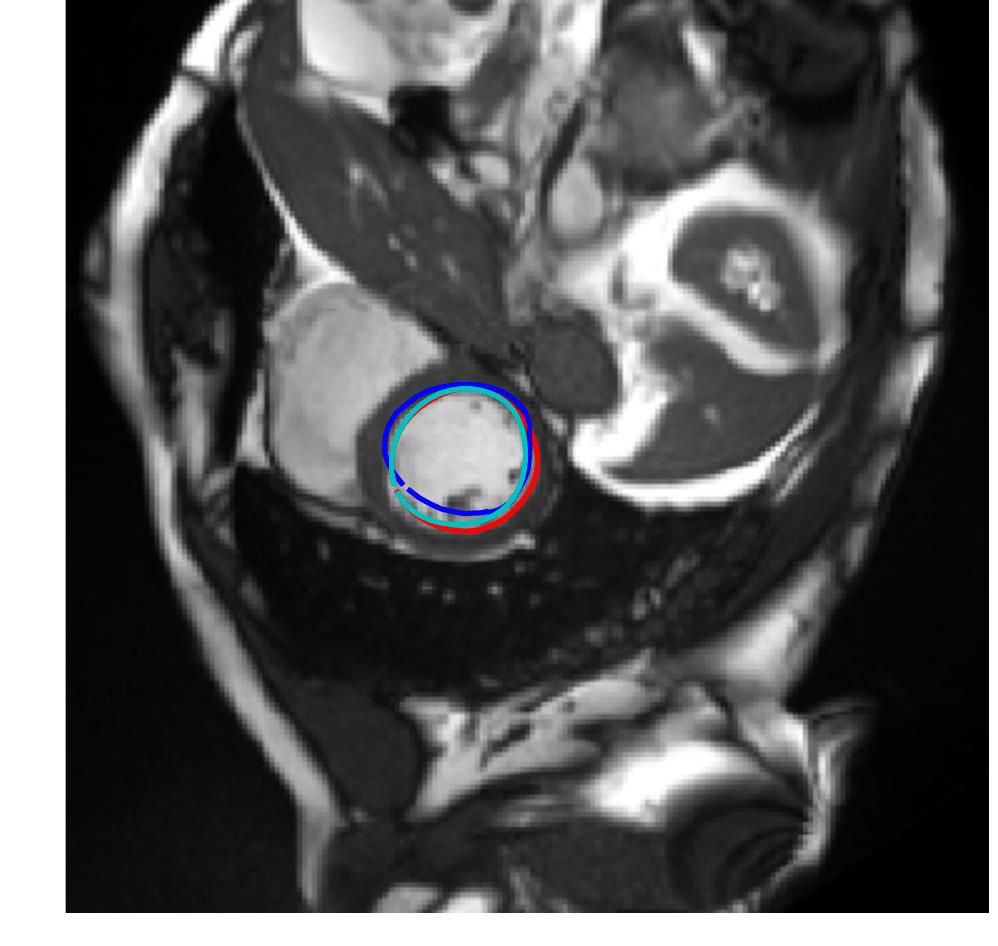}
\end{minipage}
\begin{minipage}[c]{0.325\textwidth}
	\includegraphics[angle=270,origin=c, width=\textwidth]{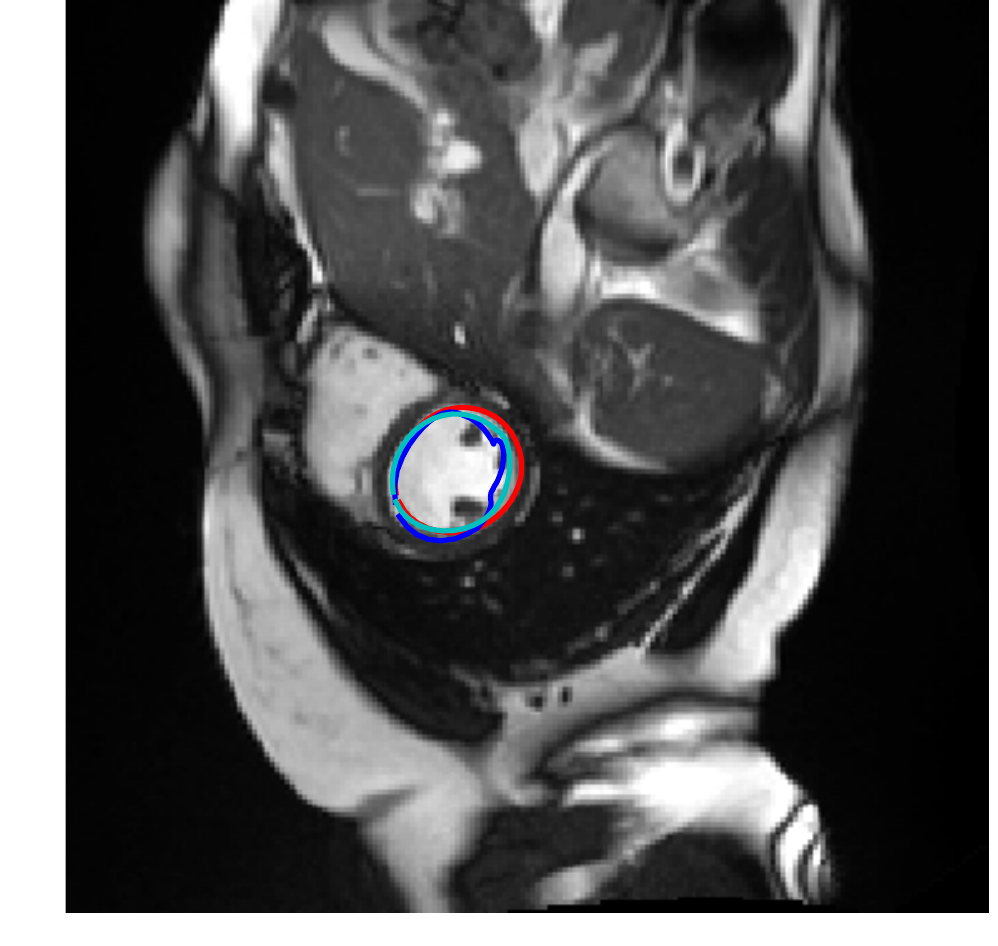}
\end{minipage}

\begin{minipage}[c]{0.325\textwidth}
	\includegraphics[angle=270,origin=c, width=\textwidth]{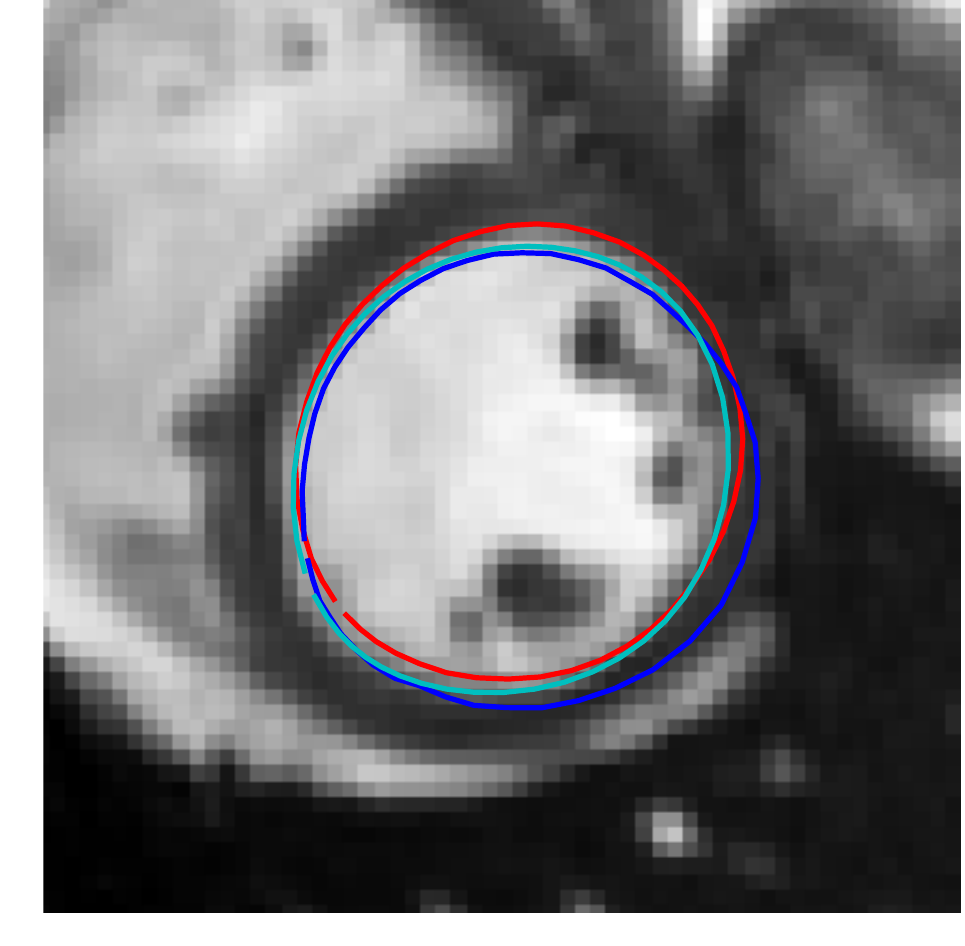}
\end{minipage}
\begin{minipage}[c]{0.325\textwidth}
	\includegraphics[angle=270,origin=c, width=\textwidth]{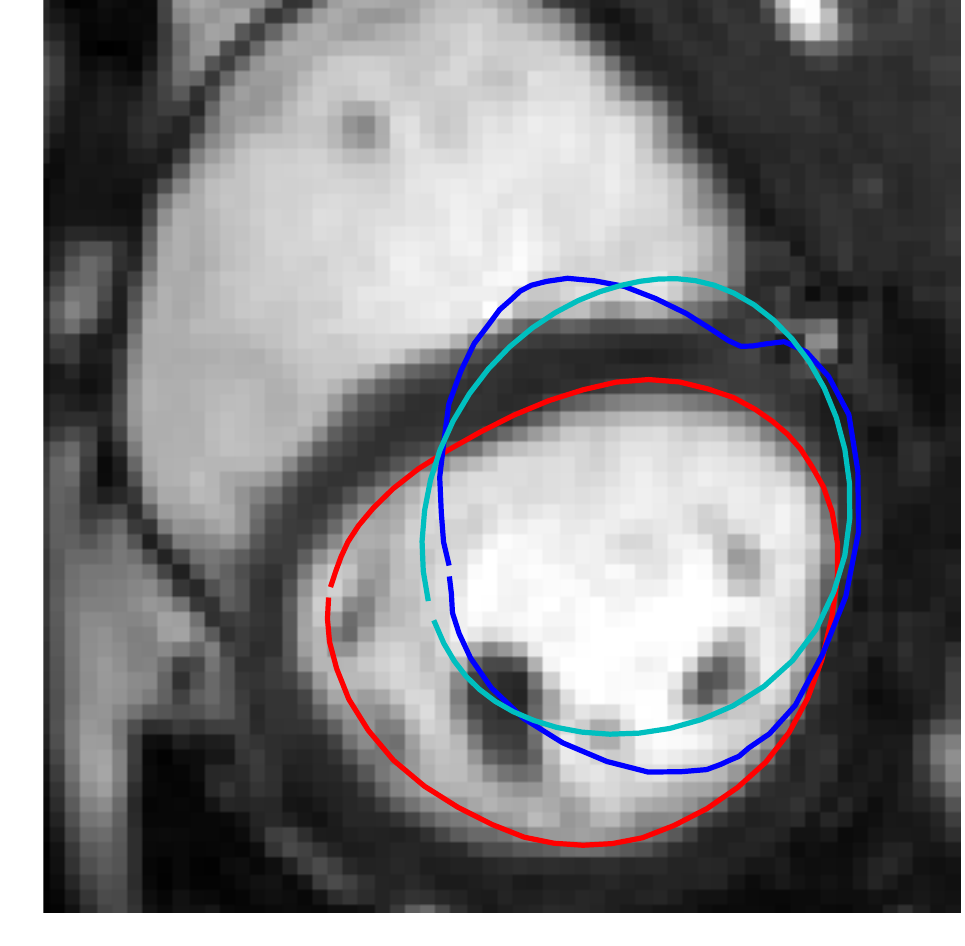}
\end{minipage}
\begin{minipage}[c]{0.325\textwidth}
	\includegraphics[angle=270,origin=c, width=\textwidth]{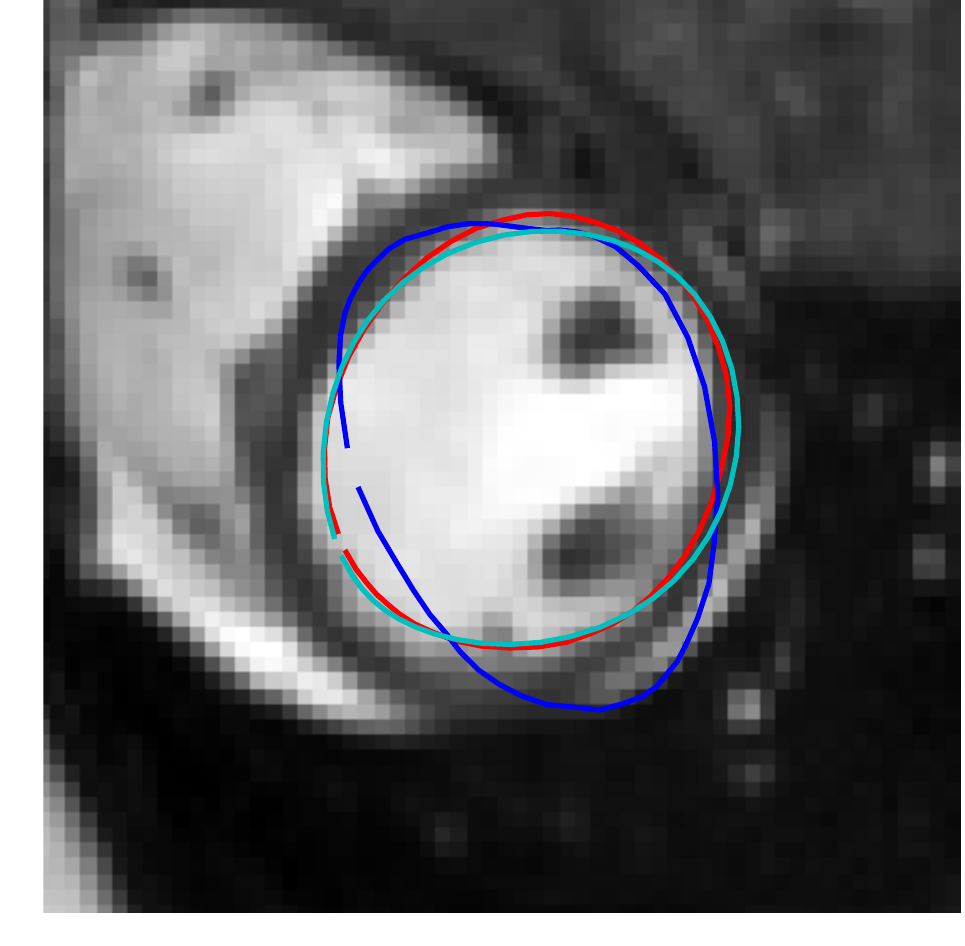}
\end{minipage}
\caption{Myocardium delineation: reference (red), predicted mean(cyan), sampled delineation (blue). We predicted positions of 50 vertices on the outline of myocardium using 12 principal components at the noise level of $5\times 10^{-2}$.Top: Model trained on full FOV images. Bottom: Model trained on small FOV images.} \label{fig:fullFOV}
\end{figure}

Finally, Figure \ref{fig:fullFOV} illustrates results of the sampling process on full and small FOV images. Bottom row images here correspond to the cases outlined in Figure \ref{smallFOV}. Observe how the samples become more variable as the uncertainty grows through subjects.

\section{Conclusion}
In this paper, we presented a novel probabilistic deep learning approach for simultaneous surface reconstruction and aleatoric uncertainty prediction. Inspired by the works on deterministic shape models \cite{milletari2017} and probabilistic PCA \cite{bishop}, our surface reconstruction method incorporates prior shape knowledge via a linear PCA model. Experiments using the UK Biobank data have shown that our probabilistic approach outperforms an analogous deterministic PCA-based method. In contrast to deterministic approaches, which provide a single surface, our method yields a distribution of positions for every vertex on the surface. This way, we can not only sample numerous predictions from one model, but ascertain the vital uncertainty pertinent to each prediction. 

Providing uncertainty estimations is essential in the medical domain, and in particular for surgery planning where precision and accuracy are of utmost importance. While the proposed method is capable of producing surface predictions of healthy 2D cardiac data, we intend to extend it to more challenging scenarios in the future. The aim is to generalise to 3D setting and other types of organs as well. Extensions to organs of more variable shapes may require adaptation of the shape model. Finally, transfer learning or domain adaptation techniques could be investigated to apply the proposed approach to datasets of substantially smaller size.

\section*{Acknowledgements}
This research has been conducted using the UK Biobank Resource under Application Number 17806.




\section*{Appendix A}
\subsection{Posterior}
From (\ref{pyz}) 
\begin{equation}
    p(y|z,x) = \mathcal{N} ( y | \underbrace{U S^{\frac{1}{2}}z + \mu + s(x)}_{:= A}, \underbrace{{\sigma}^2 I}_{:=\Sigma}), 
\end{equation}
and Jensen's inequality we have
\begin{align}
    \ln p(y|x) \geq \int \ln [p(y|z)] p(z|x) dz & = \mathbb{E}_{z|x}\left[ \ln p(y|z) \right] \\
                                                & \cong \frac{1}{L} \sum_{l=1}^L \ln p(y|z_l),
\end{align}
where $z_l$ is sampled from $p(z|x) = \mathcal{N}(z|\mu(x), \Sigma(x))$ and $\mu(x)$, $\Sigma(x)$ are provided by the network. In detail this translates to
\begin{equation}
    \frac{1}{L} \sum_{l=1}^L \ln p(y|z_l) = \frac{1}{L} \sum_{l=1}^L -\frac{1}{2} \left\{ r  \ln (2\pi) + \ln |\Sigma| + (y-A)^T \Sigma^{-1}(y-A) \right\},
\end{equation}
$r$ is the dimensionality of the vector $y$. In practice this value is summed over a batch of input vectors $x_n$. Note that to sample from $p(z|x)$ we employ the so-called "reparametrisation trick" \cite{diederik2014}, where we first sample $\epsilon \sim \mathcal{N}(0,I)$ and then compute $z_l = \mu(x) + \Sigma^{1/2}(x) \ast \epsilon$.

\subsection{Regularisation}
Considering 
\begin{align}
    p(z) &= \int p(z|x) p(x) dx \\
         & \cong \sum p(z|x_n), \qquad x_n \sim p(x),
\end{align}
the unitary Gaussian prior on $z$ and the fact that 
\begin{align}
    \mathrm{KLD}(p||q) &= \mathbb{E}_p \left[ \ln \frac{p(z)}{q(z)} \right] \\
                      & \cong \sum_{l}(\ln p(z_l) - ln q(z_l)), \qquad z_l \sim p
\end{align}
we write
\begin{align}
    \mathrm{KLD} \left( \sum_{n=1}^{N} p(z|x_n), p(z) \right) &\cong \frac{1}{L}\sum_{l}^{L}\left\{\underbrace{\ln\left[\sum_{n=1}^{N} p(z_l|x_n)\right]}_{:=LNP}\underbrace{- \ln[\mathcal{N}(0, I)]  }_{:=LNQ}\right\}
\end{align}

where $\mu_n = \mu(x_n)$ and $\Sigma_n = \Sigma(x_n)$.
Furthermore,

\begin{align}
LNP &= \ln \left[ \sum_{n} \frac{1}{\sqrt{\left( (2\pi)^s |\Sigma_n|\right)}} e^{-\frac{1}{2}(z_l - \mu_n)^T \Sigma_n^{-1}(z_l-\mu_n)}    \right] 
\end{align}

and
\begin{equation}
LNQ = - \left[ -\frac{s}{2} \ln(2 \pi) - \frac{z^Tz}{2} \right] = \frac{1}{2}\left[s \ln(2\pi) + z^T z \right],
\end{equation}
with $s$ equal to the dimensionality of the latent space.

\section*{Appendix B}
We obtain the posterior distribution of our vertex coordinates as follows.
Given
\begin{equation}\label{pyyz}
    p(y|z,x) = \mathcal{N} ( y | U S^{\frac{1}{2}} z + \mu + s(x), {\sigma}^2 I), 
\end{equation}
and
\begin{equation}
    p(z|x) = p(z|\mu(x), \Sigma(x)),
\end{equation}
we can write
\begin{align*}
    \mathbb{E}(y|x) &= \mathbb{E}(\mathbb{E}(y|z)|x) \\
                  &= \mathbb{E}(US^{\frac{1}{2}}z+\mu+s|x) \\
                  &=\int (US^{\frac{1}{2}}z+\mu+s) p(z|x) dz \\
                  &= U S^{\frac{1}{2}} \mathbb{E}(z|x) + \mu + s \\
                  &= U S^{\frac{1}{2}} \mu(x) + \mu + s. \\
\end{align*}

As for the variance,
\begin{align*}
    \mathrm{var}(y|x) &= \mathbb{E}(\mathrm{var}(y|z)|x) + \mathrm{var}(\mathbb{E}(y|z)|x) \\
                    &= \sigma^2I + \mathrm{var}(US^{\frac{1}{2}} z+\mu+s|x) \\
                    &= \sigma^2I + \mathrm{var}(US^{\frac{1}{2}} z|x) \\
                    &= \sigma^2I + U S^{\frac{1}{2}} \mathrm{var}(z|x)(US^{\frac{1}{2}})^T \\
                    &= \sigma^2I + U S^{\frac{1}{2}} \Sigma(x) (US^{\frac{1}{2}})^T.
\end{align*}

\end{document}